\documentclass[11pt]{article}

\usepackage[preprint]{acl}

\usepackage{times}
\usepackage{latexsym}
\usepackage{dsfont}
\usepackage{multirow}
\usepackage{booktabs}
\usepackage[T1]{fontenc}

\usepackage[utf8]{inputenc}

\usepackage{microtype}

\usepackage{inconsolata}

\usepackage{graphicx}
\usepackage{hyperref}
\usepackage{url}
\usepackage{algorithm}
\usepackage{algpseudocode}
\usepackage{amsmath}

\usepackage{subcaption}
\usepackage[most]{tcolorbox}
\usepackage{graphicx}
\usepackage{xcolor}
\usepackage{lipsum}

\usepackage{CJKutf8} 

\title{GraphGhost: Tracing Structures Behind Large Language Models}

\author{Xinnan Dai$^{1}$, XianXuan Long$^{1}$, Chung-Hsiang Lo$^2$,  Kai Guo$^1$\thanks{Correspondence to guokai1@msu.edu. \\The code and data are available at \\\href{https://github.com/DDigimon/GraphGhost}{\texttt{https://github.com/DDigimon/GraphGhost}}},\\ \textbf{Shenglai Zeng}$^1$, 
\textbf{Dongsheng Luo}$^{3}$, \textbf{Jiliang Tang}$^{1}$ 
\\ 
$^1$Michigan State University 
\quad $^2$ Northeastern University\\
\quad $^3$ Florida International University
  \\
\{daixinna, guokai1\}@msu.edu
}

\begin{document}
\maketitle
\begin{abstract}
Large Language Models (LLMs) exhibit strong reasoning capabilities on structured tasks, yet the internal mechanisms underlying such behaviors remain poorly understood. Existing interpretation methods mainly focus on token-level attributions, which provide limited insight into multi-step reasoning inside the model.
We propose GraphGhost, a graph-based framework that models internal token interactions and neuron activations in LLMs as graphs. By aggregating token dependencies traced across layers, GraphGhost captures global information flow underlying model predictions. We formalize GraphGhost from two complementary perspectives: a sample view, which traces token dependencies for individual predictions, and a dataset view, which aggregates recurring structural patterns learned during training.
Through graph analytics and quantitative experiments, we show that graph structural properties are closely associated with influential tokens and neuron nodes, and that perturbations to structurally critical nodes lead to measurable changes in reasoning behavior.
These results indicate that the structural patterns captured by GraphGhost reflect meaningful internal organization of LLM reasoning. The codes are available at software part. Artifacts will be made available for research use only.
\end{abstract}
\section{Introduction}


Large Language Models (LLMs) have demonstrated remarkable capabilities in solving structured reasoning tasks, 
where solutions require the composition of multiple intermediate steps governed by explicit relational or logical dependencies, 
such as mathematical problem solving, logical inference, and graph-based reasoning~\citep{bommasani2022opportunitiesrisksfoundationmodels, pan2023logiclmempoweringlargelanguage}. 
Despite this success, the internal mechanisms by which LLMs represent, compose, and reuse such structural information during reasoning remain poorly understood~\citep{nanda2023progressmeasuresgrokkingmechanistic, nananukul2025logicalthoughtlogicbasedontologicalgrounding, pandey2025adaptivegraphthoughtstesttime}. This is important because it provides a principled way to understand the source of intelligence in LLMs.


\begin{figure}

\vspace{-12pt}
    \centering
    \includegraphics[width=1\linewidth]{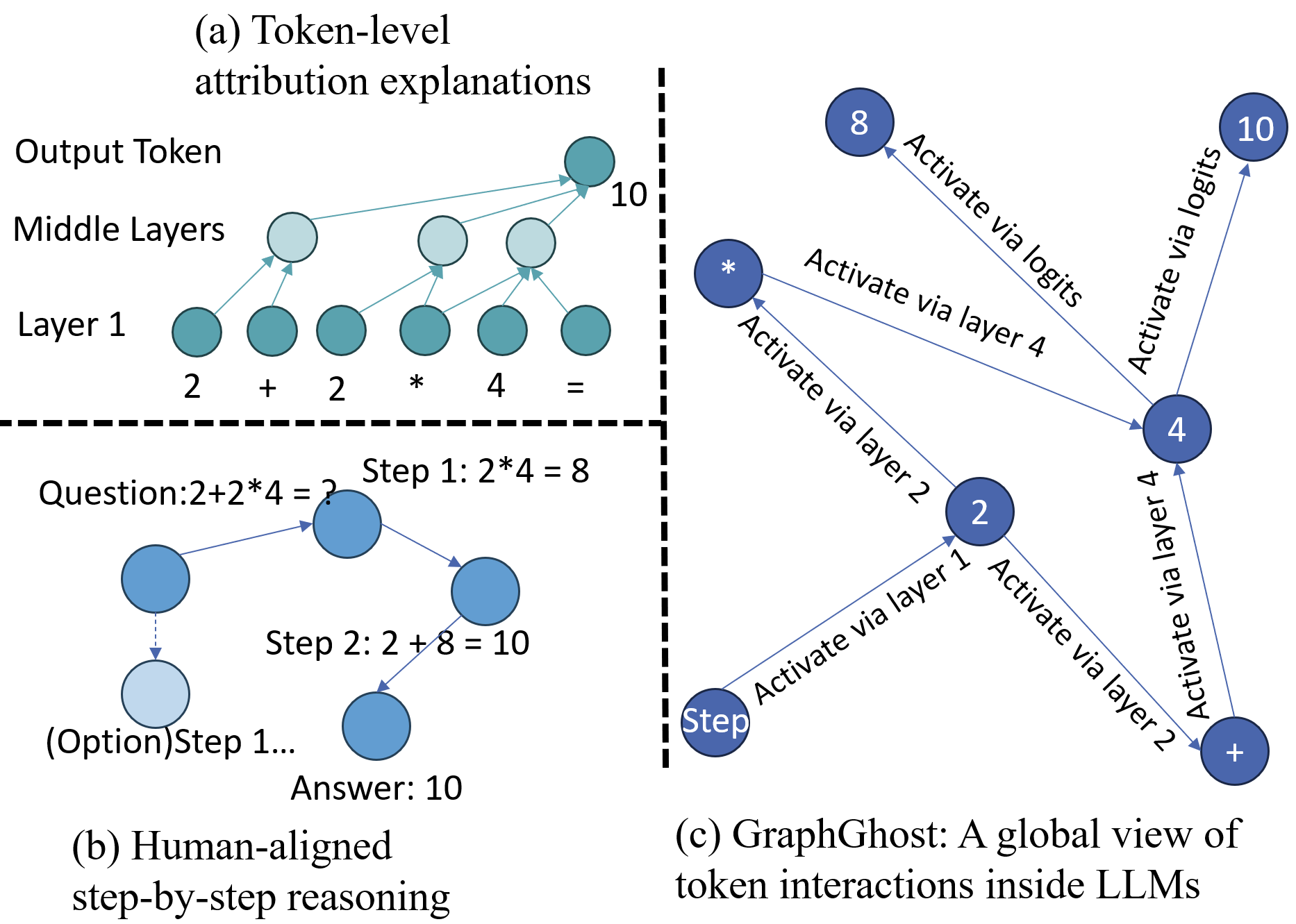}
    \caption{From local explanations and human-aligned reasoning to global internal structures. (a) Token-level attribution explanations provide localized views of individual predictions.
(b) Human-aligned step-by-step reasoning produces coherent reasoning traces without revealing internal mechanisms.
(c) GraphGhost captures global token interactions inside LLMs as a structured graph, enabling mechanistic analysis of internal reasoning processes.}
    \label{fig:intro}
    \vspace{-12pt}
\end{figure}

Most existing efforts to understand LLMs' behaviors on structured reasoning tasks focus on token-level explanations~\citep{chen2025doeschainthoughtthink, tak2025mechanisticinterpretabilityemotioninference}, as illustrated in Figure~\ref{fig:intro}(a),
which trace individual predictions back to contributing tokens or activations and provide only localized views of isolated outputs.
Given that reasoning capability plays a central role in LLM performance across reasoning tasks~\citep{wei2022chain, lewkowycz2022solvingquantitativereasoningproblems} 
as shown in Figure~\ref{fig:intro} (b), existing token-level approaches provide limited insights into how representations are composed and reused across multiple reasoning steps~\citep{geiger2025causalabstractiontheoreticalfoundation, dziri2023faithfatelimitstransformers}. This limitation is increasingly recognized as a major obstacle to meaningful interpretability, as faithful understanding requires alignment between internal reasoning mechanisms and human-interpretable reasoning logic~\citep{zhao2024explainability, turpin2023language, barez2025chain}.

To bridge this gap, we introduce GraphGhost, a graph-based framework that captures internal token interactions and neuron activations in LLMs as graphs as shown in Figure~\ref{fig:intro}(c). Motivated by the central role of step-by-step reasoning in LLM performance, GraphGhost moves beyond isolated token attributions to explore how the reasoning steps are driven under the individual token explanation. This structural representation enables us to analyze reasoning as an internal, mechanistic process, revealing how intermediate tokens are merged, propagated, and reused during generation.

Specifically, in Section~\ref{sec:graphghosts}, we formalize GraphGhost from two complementary perspectives. First, in the sample view, GraphGhost traces internal token dependencies within individual input–output pairs that contribute to a specific prediction, exposing reasoning structures for LLMs. 
Second, in the dataset view, GraphGhost is constructed by aggregating structures across samples to reveal recurring and shared token-level patterns learned from data. Through graph analytics, we demonstrate that LLM reasoning behavior is tightly coupled to internal graph structural properties, with influential tokens and neuron nodes characterized by structurally critical roles.
These findings suggest that internal graph structures fundamentally shape how individual tokens influence LLM reasoning, through their structural roles within the model.


In Section~\ref{sec:controlling}, we conduct extensive quantitative experiments to assess whether the structural patterns captured by internal graphs are meaningfully tied to LLM reasoning behavior.
Using fidelity, infidelity, and sparsity metrics, we demonstrate that the tokens and neuron nodes identified by GraphGhost exert a significantly strong influence on model behavior. Furthermore, targeted interventions on structurally important nodes lead to measurable changes in reasoning trajectories and outputs, providing empirical evidence that GraphGhost captures meaningful internal structures governing LLM reasoning.

In conclusion, we summarize our key contributions as follows:

1) We propose GraphGhost, a unified framework for understanding how LLMs perform reasoning by modeling internal token interactions and neuron activations at both the sample and dataset levels.
    

2) Using GraphGhost, we reveal that internal graphs capture structural constraints that govern model behavior, shaping information composition and propagation during reasoning.

3) We conduct quantitative experiments to test whether graphs constructed by GraphGhost capture meaningful structural patterns of LLM reasoning. Our results show that structurally important token-related neuron nodes strongly influence reasoning trajectories, and that intervening on these nodes alters both language expression and logical flow.

\section{Related Work}


While Chain-of-Thought (CoT) prompting enhances LLM reasoning~\citep{wei2022chain, kojima2022large}, its faithfulness to the model’s intrinsic computations remains contested~\citep{jacovi2020faithfullyinterpretablenlpsystems, lyu2023faithfulchainofthoughtreasoning}. Prior work shows that CoT often acts as post-hoc rationalization, generating explanations that diverges from the underlying causal logic~\citep{turpin2023language} or internal attention patterns~\citep{lanham2023measuringfaithfulnesschainofthoughtreasoning, barez2025chain}. Moreover, such behavioral analyses provide limited insight into the specific neurons or circuits driving model predictions.

To overcome the lack of causal rigor in probing~\cite{alain2018understandingintermediatelayersusing, belinkov2021probingclassifierspromisesshortcomings} and attention analysis~\cite{clark2019doesbertlookat, vig2019analyzingstructureattentiontransformer}, research has pivoted to attribution methods~\citep{sundararajan2017axiomaticattributiondeepnetworks, meng2023locatingeditingfactualassociations} and circuit tracing~\citep{elhage2021mathematical}. Recent methods leverages Sparse Autoencoders (SAEs)~\cite{bricken2023towards, cunningham2023sparseautoencodershighlyinterpretable} and Transcoders~\cite{dunefsky2024transcoders} to decompose dense representations into interpretable features~\citep{bloom2024saetrainingcodebase, ferrando2025iknowentityknowledge, paulo2025automaticallyinterpretingmillionsfeatures}, enabling attributed graphs~\citep{circuit-tracer, marks2025sparsefeaturecircuitsdiscovering} to model causal flow. GraphGhost extends this framework to uncover internal structures that affect LLM behavior in both sample and dataset view.

\begin{figure*}
\vspace{-30pt}
    \centering
    \includegraphics[width=1\linewidth]{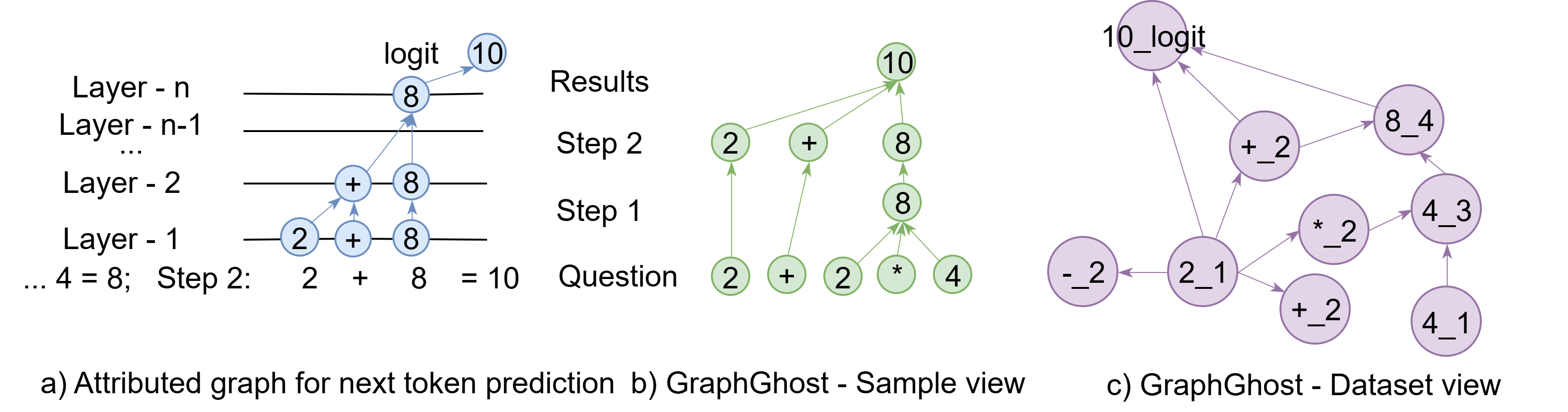}
    \caption{We use a simple example: 2 + 2 * 4.
(a) An attributed graph interpreting a single-token prediction.
(b) GraphGhost (sample view) aggregates token dependencies along the reasoning chain.
(c) GraphGhost (dataset view) captures cross-layer token correlations learned from the corpus.
Here, ``2\_1'' denotes token 2 at layer 1; ``-\_2'' does not appear in the sample but is activated from other corpus contexts.
Token ``2\_1'' participates in both ``2*4'' and ``2+8'' computations.}
    \label{fig:method}
    \vspace{-12pt}
\end{figure*}

\section{Tracing the GraphGhosts}
\label{sec:graphghosts}

The reasoning behavior of LLMs emerges through successive token generations.
Accordingly, existing interpretation methods primarily operate at the token level, attributing individual predictions to specific tokens or activations and providing localized explanations, as illustrated in Figure~\ref{fig:method}(a)~\cite{ameisen2025circuit}.
However, reasoning does not arise from isolated tokens alone, but from structured interactions among multiple tokens within a sample and from recurring patterns shared across samples.

To capture these aspects, we introduce two complementary views: a sample view and a dataset view.
The sample view models token interactions within a single reasoning instance and formalizes this instance-specific structure as a GraphGhost (Figure~\ref{fig:method}(b); Section~\ref{sec:graphghost_sample}).
The dataset view aggregates such structures across samples to reveal reusable reasoning patterns and is formalized in Section~\ref{sec:graphghost_dataset} (Figure~\ref{fig:method}(c)).
\vspace{-10pt}
\subsection{Sample View}
\label{sec:graphghost_sample}

\begin{figure}[htbp]
\vspace{-25pt}
    \centering
    \begin{subfigure}[b]{0.2\textwidth}
        \centering
        \includegraphics[width=\linewidth]{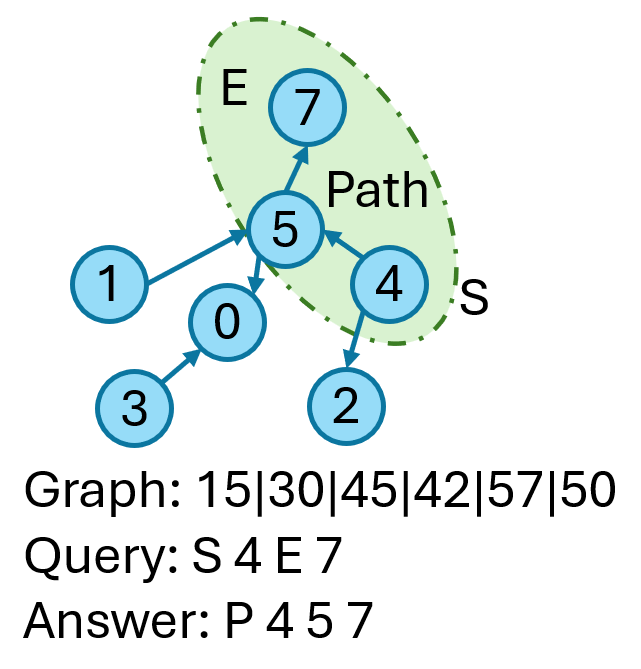}
        \caption{Graph description in a simplified way. The sequence includes the graph description, query, and answer.}
        \label{fig:sync_graph}
    \end{subfigure}
    \hfill
    \begin{subfigure}[b]{0.25\textwidth}
        \centering
        \includegraphics[width=\linewidth]{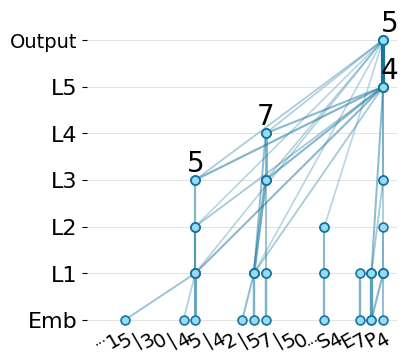}
        \caption{Interpretation of the graph path reasoning task. The bolder lines indicate stronger contributions to neuron activations.}
        \label{fig:sync_vis}
    \end{subfigure}
    \caption{A toy example of path reasoning on a graph. (a) The simplified descriptions to predict the path between 4 and 7; (b) The circuit tracing for the interpretation of why token 5 is selected to be predicted. }
    \label{fig:sync_all}
    \vspace{-14pt}
\end{figure}

LLMs have demonstrated strong capabilities in chain-of-thought reasoning.
However, whether these reasoning chains faithfully correspond to coherent internal reasoning mechanisms remains unclear.
Rather than attempting to answer this question conclusively, we begin with a toy example to illustrate how circuit-tracing methods can be used to probe the internal computations underlying such reasoning behaviors.
\vspace{-10pt}
\paragraph{Toy Example on Graph Reasoning}
Since search paths on a graph naturally decompose reasoning into node-by-node transitions, we adopt the path-reasoning task as a toy example to illustrate how nodes are selected at each step of the reasoning process. Following prior work~\cite{saparov2025transformersstrugglelearnsearch, wang2024alpineunveilingplanningcapability} that studies LLM reasoning and planning through simplified graph abstractions, we focus on shortest-path search as our canonical setting, as illustrated in Figure~\ref{fig:sync_graph}.

Graphs are specified as lists of edges, with the symbol ``|'' used to separate individual edges. In the queries, we denote the start and end nodes using the markers ``S'' and ``E'', respectively. In the answers, we use the symbol ``P'' to indicate that the model is required to generate an explicit reasoning path. we train a 5-layer GPT-2 architecture decoder-only transformer on this task. The trained model achieves 98\% accuracy on the test set.
To understand how the trained model internally performs path finding, we analyze its inference process using circuit-tracing methods.

Specifically, following~\cite{circuit-tracer}, we train a transcoder~\cite{dunefsky2024transcoders} for each layer of the transformer.
Each transcoder takes the layer-wise hidden activations as input and learns a sparse feature representation that reconstructs these activations, yielding token-level semantic features for each MLP layer.
We then replace the original MLP layers with the corresponding transcoders and compute the logit-level contribution weights of each transcoder feature to the final prediction.
Using transcoder features as nodes and their logit contribution weights as directed edges, we construct the attributed graph. The resulting attribution graph, visualized in Figure~\ref{fig:sync_vis}, illustrates how the transformer models the path-finding process.
Full experimental details are provided in Appendix~\ref{app_sec:sync_experiment}.

Circuit tracing reveals that decoder-only transformers revisit the preceding context to identify relevant nodes and predict the next token. From a structural perspective, neighboring nodes are progressively merged in the middle layers. For example, at the third layer, node 4 is merged with node 5, forming a semantic representation of the edge between them. During generation, the final token in the sequence receives the signal that an edge exists between nodes 4 and 5 at layer 3, which leads the model to predict node 5 as the next token. 
This mechanism differs from human graph reasoning, which explicitly retrieves relations, checks connectivity, and composes shortest paths. We then evaluate this assumption on real LLM reasoning tasks.
\vspace{-20pt}
\paragraph{GraphGhost in Sample Interpretation}

In the toy example, we observe that reasoning in basic decoder-only transformers is driven by token-level associations, 
where previously observed in-context tokens contribute to information aggregation during token generation, 
rather than through explicitly modeled logical structures.
Motivated by this observation, we merge answer-related circuits to examine how reasoning is internally realized for a specific query in LLMs.


At the sample level, GraphGhost aims to capture the internal structural dependencies that contribute to a model’s final prediction.
Given a sample $s$ with final answer token $t_n$, we initialize an empty graph $G^{\text{sample}} = (V, E)$ together with a local edge-weight accumulator $W^{\text{local}}$,
which is used to aggregate contribution strengths when the same dependency is discovered multiple times during backward expansion.
The graph construction starts from $t_n$ and iteratively expands backward through the model’s internal computation.

For each token $t$ in the current frontier that is not part of the question tokens, we apply a circuit tracing procedure to extract 
the local logit-based attribution graph rooted at $t$ follows the previous paragraph. This step yields a set of contributing tokens as sample-view nodes $\hat{V}$, directed edges $\hat{E}$, and local attribution weights $W^{\text{loc}}$, characterizing how information flows to the prediction of $t$. Newly discovered tokens are added to the frontier to enable recursive exploration of upstream dependencies.

The nodes and edges from each local attribution graph are 
by taking the union over nodes and edges, removing duplicates and retaining only the existence of dependency relations. As a result, the sample-level GraphGhost is an unweighted structural graph that captures which token-level dependencies are present, independent of their individual contribution strengths. In parallel, local edge weights are accumulated in $W^{\text{local}}$ for each observed edge, which is retained solely for later aggregation at the dataset level. The final output is a sample-specific GraphGhost graph $G^{\text{sample}} = (V, E)$ together with the local edge-weight map $W^{\text{local}}$. The overall algorithm is shown in Algorithm~\ref{alg:graphghost_sample} in Appendix~\ref{app_sec:alg}.

\label{sub_sec:case_study_graphghost}
\begin{figure}
\vspace{-25pt}
    \centering
    \includegraphics[width=\linewidth]{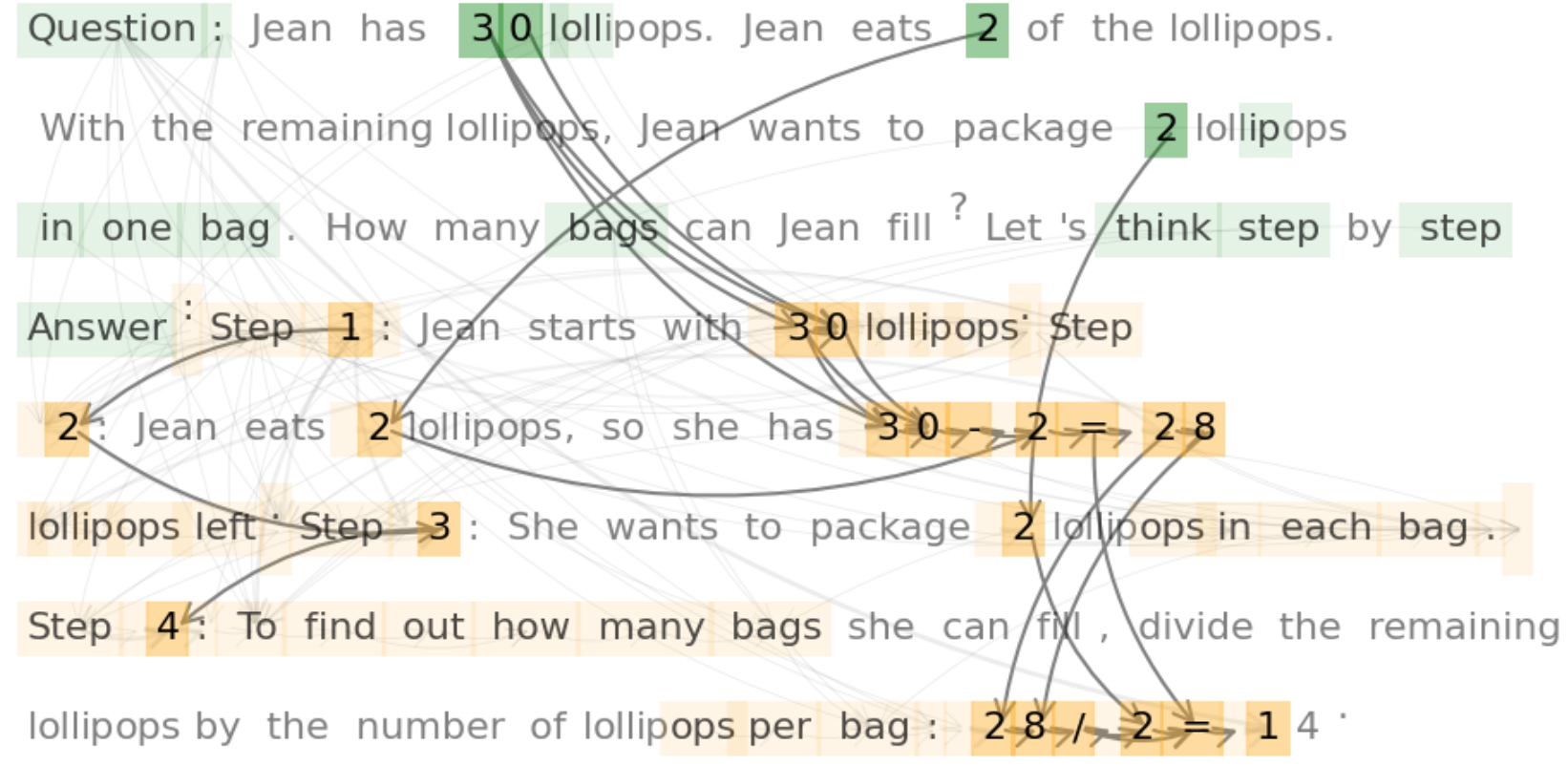}
    
    \caption{Case study tracing tokens contributing to the answer 14 in Qwen3-0.6B, illustrating how GraphGhost captures the numerical reasoning flow linking number-related tokens to the final prediction.}  
     
    \label{fig:intrasample_graph}
    \vspace{-16pt}
\end{figure}


The sample view of GraphGhost characterizes step-by-step reasoning in LLMs by modeling structural dependencies among tokens as reasoning unfolds. Here, we show an example in Figure~\ref{fig:intrasample_graph}. The identified token set $T$ includes tokens from both the input question (highlighted in green) and those generated along the reasoning chain (in orange). 
As shown in the figure, number-related tokens emerge through three distinct mechanisms: 
(i) step-by-step progression (e.g., 1–2–3–4), 
(ii) direct copying from the input context (e.g., 30 and 28), and 
(iii) generation influenced by neighboring tokens, such as the digit “1” in the final answer “14”, which is activated by ``=". 
This tracing process demonstrates how local dependencies between tokens collectively form reasoning paths during sequential generation.




\vspace{-5pt}
\subsection{Dataset View}
\label{sec:graphghost_dataset}
The reasoning capabilities of LLMs are learned from large-scale corpora, which motivates the assumption that certain neuron units within LLMs encode reusable token-level connections that support generalization to new reasoning tasks. Accordingly, we introduce the dataset-level view of GraphGhost to explore the implicit structures inside LLMs.
\vspace{-5pt}
\subsubsection{Method}
To build the dataset-level of GraphGhost, we aggregate sample-level reasoning structures to reveal recurring internal information-flow patterns shared across multiple samples from datasets.
Given a set of samples $\mathcal{S} = \{s_1, \ldots, s_m\}$, we construct a unified graph for the dataset as
$G^{\text{data}} = (V^{\text{data}}, E^{\text{data}}, W^{\text{data}})$.


For each sample $s_i \in \mathcal{S}$, we first obtain its sample-level GraphGhost
$G^{(i)} = (V^{(i)}, E^{(i)})$ together with a local edge-weight map $W^{\text{local}}$
using the sample-view construction.
The dataset-level node and edge sets are formed by merging sample graphs via set union:
$V^{\text{data}} = \bigcup_i V^{(i)}$ and $E^{\text{data}} = \bigcup_i E^{(i)}$.
In parallel, edge attribution strengths are accumulated across samples.
For each edge $e \in E^{(i)}$, we update its dataset-level weight as
$
W^{\text{data}}(e) \leftarrow W^{\text{data}}(e) + W^{\text{local}}(e),
$
where $W^{\text{local}}$ captures unnormalized, sample-specific contribution strengths.
This accumulation preserves both the frequency and strength of recurring dependencies across samples.
After aggregating all samples, we apply row-stochastic normalization to $W^{\text{data}}$
to obtain relative edge importance at the dataset level.
The resulting graph captures recurring reasoning paths,
while the normalized weights quantify their global importance across the dataset.
The full procedure is summarized in Algorithm~\ref{alg:graphghost_dataset} in Appendix~\ref{app_sec:alg}.

\begin{figure}
\vspace{-25pt}
    \centering
    \includegraphics[width=0.8\linewidth]{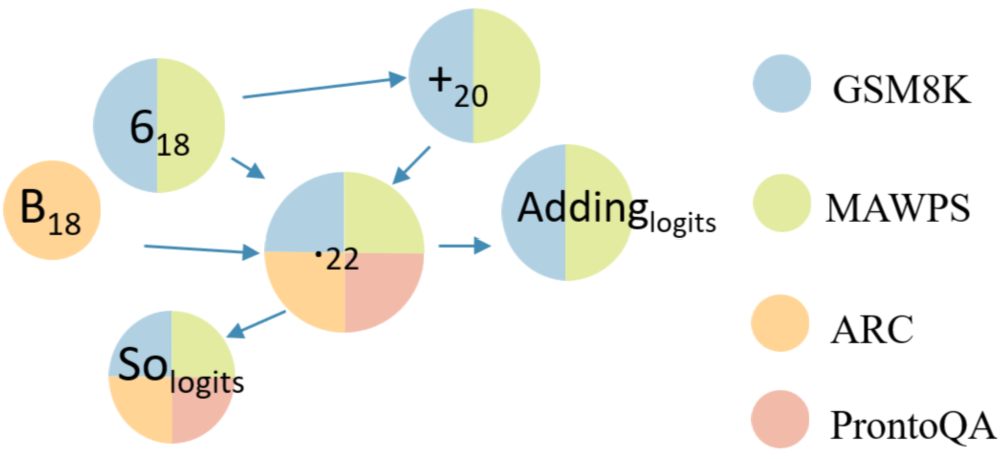}
    \caption{A subgraph of GraphGhost in Qwen3-0.6B model. We include 4 datasets from different domains.}
    \label{fig:token_level_graph}
    \vspace{-12pt}
\end{figure}

The dataset view aims to reveal whether LLMs exhibit recurring internal structures across different tasks and datasets.
To this end, we visualize a subgraph extracted from the dataset-level GraphGhost in Figure~\ref{fig:token_level_graph},
which aggregates sample-level GraphGhosts constructed from four datasets spanning different domains.
In this graph, each node represents a token–layer pair identified by GraphGhost,
and node colors indicate the datasets in which the token participates, while mixed colors denote tokens shared across multiple datasets. From this visualization, several consistent patterns emerge. First, certain tokens appear across all four datasets, suggesting shared internal mechanisms.
For example, the period symbol ``.'' at layer 22 exhibits strong logit contributions across all datasets,
and the discourse marker ``So'' frequently appears in diverse reasoning chains. Second, domain-specific patterns are also evident: arithmetic-related tokens such as ``Adding'' and expressions like ``6+'' are shared between GSM8K and MAWPS, which both involve numerical reasoning.
In contrast, some nodes are task-specific; for example, $\mathrm{B_{18}}$ appears only in ARC, reflecting choice-based reasoning. This shows that dataset-level GraphGhost captures both shared reasoning primitives and task-specific structures.

\vspace{-5pt}
\subsubsection{Analysis}
\label{sec:graphghost_analysis}

Beyond visualization, we conduct further graph-based analyses to examine how the structures identified by GraphGhost correspond to LLM behaviors.
Specifically, we analyze the structures from two perspectives:
(1) token-merging behavior, quantified by node in-degree; and
(2) trigger tokens that initiate reasoning, identified using PageRank.
In addition, we present case studies to illustrate how perturbations based on GraphGhost-identified structures affect the progression of the reasoning process.


We consider Qwen3-0.6B (28 layers), Qwen2.5-1.5B (28 layers), Llama3.1-8B-Instruct (32 layers), DeepSeek-distilled-Qwen2.5B-1.5B (DS-Qwen-1.5B; 28 layers), and DeepSeek-distilled-Llama-8B (DS-Llama-8B; 32 layers).
We evaluate these models on three reasoning domains: (i) mathematical reasoning (GSM8K, MAWPS), (ii) logical reasoning (ProntoQA, BoolQA), and (iii) scientific reasoning (ARC-Easy, QASC). We obtain the GraphGhosts for each model and dataset. All experiments are conducted on a single H200 GPU. Each sample will run 0.5-3h, depended on the length of reasoning tokens. Full experimental details are provided in Appendix~\ref{app_sec:real_world_exp}.


\paragraph{Semantic Merging}
\vspace{-8pt}

We begin by analyzing semantic merging behavior using the average in-degree of neuron nodes across layers.
We assume that multiple tokens converge within the network, indicating potential semantic consolidation.
For instance, in the representation of the number ``10'', the tokens ``1'' and ``0'' exhibit convergent interactions in intermediate layers, consistent with the formation of a unified numerical representation.

In the GraphGhost representation, in-degree quantifies how many contributing nodes are merged into a target node.
Accordingly, nodes with higher in-degree correspond to tokens that integrate information from multiple sources
and serve as central aggregation points for reasoning-related signals.
Based on this, we examine which layers and tokens are preferentially in LLMs by GraphGhost structures.

\begin{figure}
    \centering
    \vspace{-25pt}
    \includegraphics[width=0.95\linewidth]{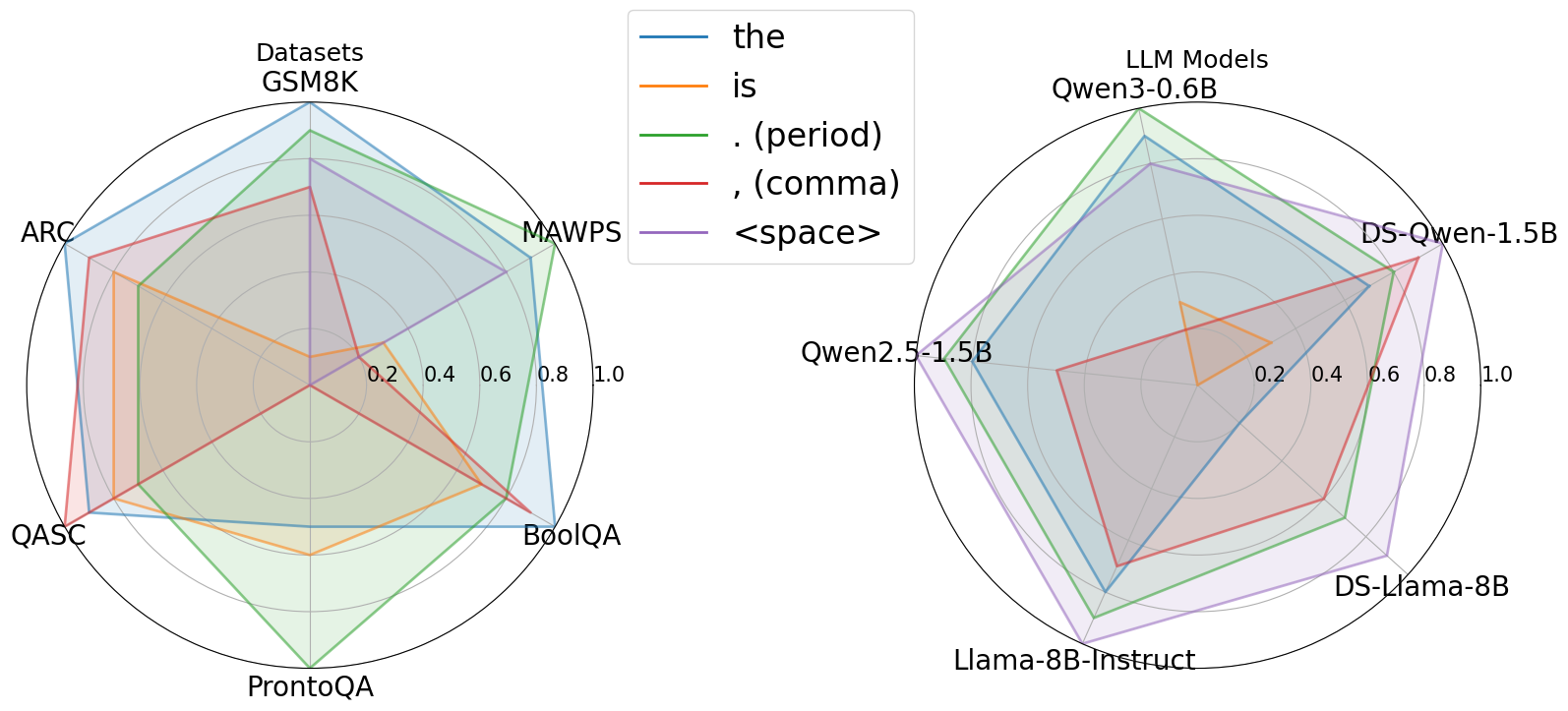}
    \caption{The top merging nodes evaluated by in-degree ratio. Left: The highest score of the in-degree ratio across datasets.
Right: The highest score of the in-degree ratio across the models}
    \label{fig:semantic_range}
    \vspace{-12pt}
\end{figure}

The statistical results are summarized in Appendix~\ref{app_sec:token_results} and visualized in Figure~\ref{fig:semantic_range}.
We observe that tokens with high in-degree exhibit clear domain sensitivity.
For example, in the ProntoQA dataset, the model is more sensitive to the period token than to the comma,
suggesting that punctuation plays different semantic roles across reasoning domains.
Across models, however, the dominant tokens are largely consistent:
special tokens such as \texttt{<space>}, the period, and ``the'' consistently rank among the top,
and this pattern persists even after finetuning.

We further set the \textit{in-degree ratio} to measure the merging behavior across the datasets and models. Let $d_{i}^{(\ell)}$ denote the in-degree of neuron node $(t_i, \ell)$, where $\ell$ is the layer index and $i$ indexes the neuron within that layer. 
The in-degree ratio of layer $\ell$ is then defined as
$
R_{\ell} = \frac{\sum_{i=1}^{N_{\ell}} d_{i}^{(\ell)}}{\sum_{k=1}^{L} \sum_{i=1}^{N_{k}} d_{i}^{(k)}},
$
where $N_{\ell}$ is the number of neurons in layer $\ell$ and $L$ is the total number of layers. 
A higher in-degree ratio $R_{\ell}$ implies that layer $\ell$ concentrates a larger fraction of merging behavior,
indicating that many reasoning-related signals converge at this layer.
In contrast, a lower ratio suggests that neurons in the layer are less involved in integrating information,
playing a more distributive or transitional role in the reasoning process.

\begin{figure}
    \centering
    \vspace{2pt}
    \includegraphics[width=0.9\linewidth]{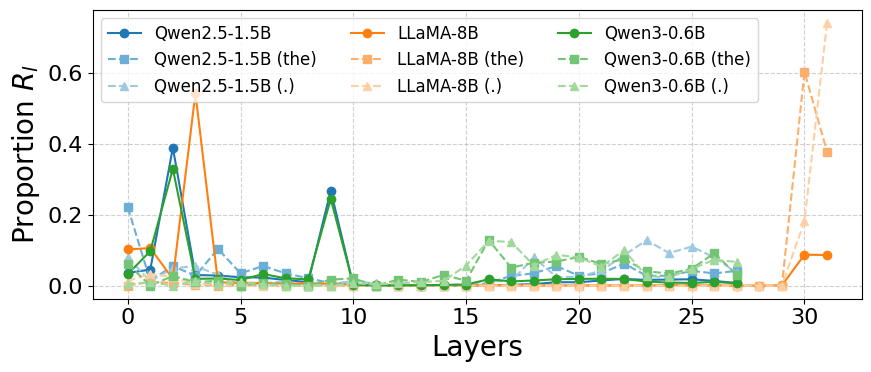}
    \vspace{-8pt}
    \caption{The in-degree proportions on various layers with MAWPS dataset}
    \label{fig:semantic_layers}
    \vspace{-12pt}
\end{figure}

As shown in Figure~\ref{fig:semantic_layers}, the in-degree ratio distribution is imbalanced across both models and tokens. Although Qwen models and the Llama model perform semantic merging primarily in the bottom and top layers, the exact layers differ between them. Similarly, token merging patterns are not always consistent. Most tokens are merged in shallow layers, whereas 
high in-degree tokens such as period and ``the" tend to merge at higher layers.

\begin{figure}
\vspace{-25pt}
    \centering
        \includegraphics[width=\linewidth]{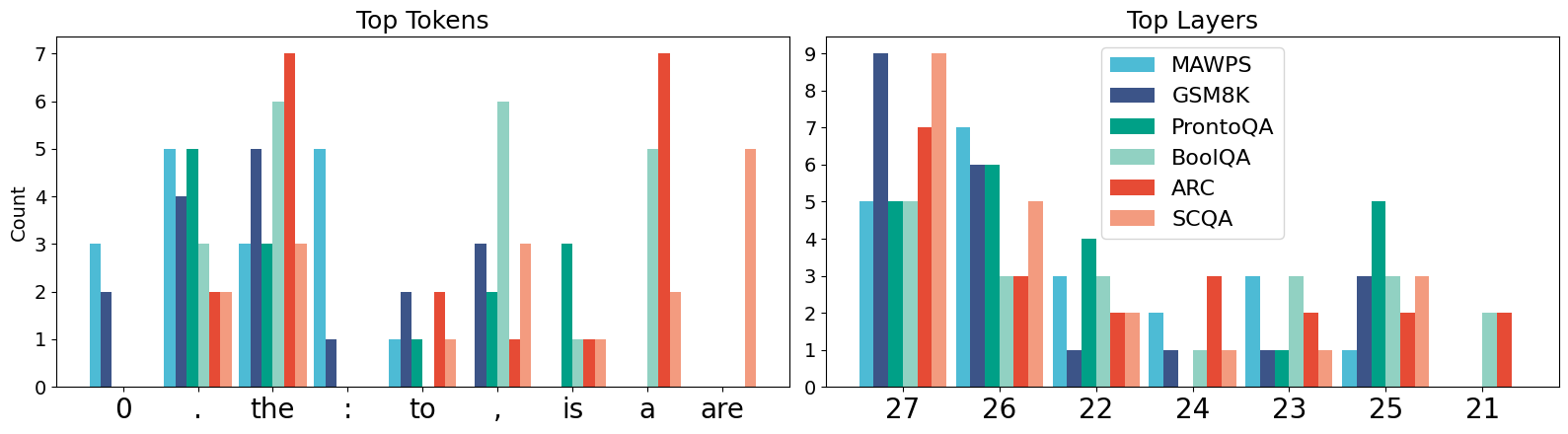}
        \caption{Top contributing tokens and layers in the dataset-level GraphGhost of Qwen3-0.6B across different datasets.
Left: The most frequently contributing tokens.
Right: The most frequently contributing tokens.}
        \label{fig:token_data}
        \vspace{-12pt}
\end{figure}

\begin{figure}
        \centering
        \includegraphics[width=\linewidth]{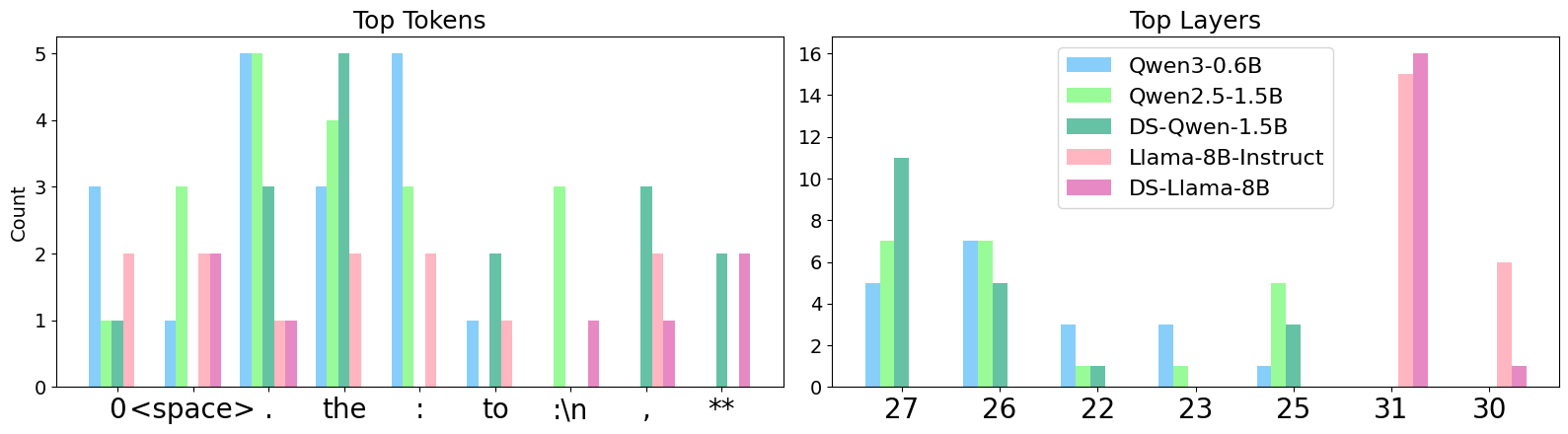}
        \caption{Top contributing tokens and layers across models evaluated by MAWPS dataset. Left: The most frequently contributing tokens.
Right: The most frequently contributing tokens.}
        \label{fig:token_model}
        \vspace{-12pt}
\end{figure}

\begin{figure*}[!t]
\vspace{-30pt}
\centering
\resizebox{0.85\textwidth}{!}{%
  \begin{minipage}{1.4\textwidth}
    \begin{tcolorbox}[title=Case studies, colback=white, colframe=black!50]
      \textbf{Semantic merging controls language}\par

      \textit{Question:} {A ) nervous and digestive B ) nervous and circulatory C ) respiratory and digestive D ) respiratory and circulatory  Question: Which two body systems would most directly remove extra fluid from a person's lungs?}\par
        
      \textit{Original answer:} {... B) Nervous and circulatory systems: The nervous system is responsible for regulating the body's response to stress, while the circulatory system is responsible for transporting blood throughout the body. Neither of these systems directly remove extra fluid from the lungs...}\par
      \textbf{\textit{Mute the neuron node 'the' at layer 20}}\par
      \textit{Perturbed answer:} {... B) Nervous and circulatory systems: The nervous system is responsible for controlling the - circulatory system is responsible for transporting blood throughout the\begin{CJK}{UTF8}{gbsn}分区肺\end{CJK}. While the\begin{CJK}{UTF8}{gbsn}分区肺需要血液来清除多余的水分，这些系统本身并不直接参与这个过程。\end{CJK}...}\par
    \noindent\rule{\linewidth}{0.2pt}
          \textbf{Reasoning trigger changes logic}\par

      \textit{Question:} {There are 7.0 crayons in the drawer and 6.0 crayons on the desk. Sam placed 4.0 crayons and 8.0 scissors on the desk. How many crayons are now there in total?
Let's think step by step (ground truth: 17)}\par
        
      \textit{Original answer:} {... - Sam placed 4.0 crayons on the desk.
- Sam also placed 8.0 scissors on the desk. .... - Total crayons on the desk = 7.0 + 4.0 + 8.0
- Total crayons on the desk = 19.0}\par
      \textbf{\textit{Mute the neuron node '.' at layer 25}}\par
      \textit{Perturbed answer:} {... - Sam placed 4.0 crayons on the desk.
- So, the total number of crayons on the desk is now 6.0 (initial crayons) + 4.0 (Sam's crayons) = 10.0 crayons. ... - So, the total number of crayons = 7.0 (drawer) + 10.0 (desk) = 17.0 crayons.}



    \end{tcolorbox}
  \end{minipage}
}
\caption{Semantic control and reasoning triggers change the model generation processing. In the language-switching case, the generated characters are direct translations of the original answer.} 
\label{fig:case_study_semantic}
\vspace{-12pt}
\end{figure*}
\paragraph{Reasoning Triggers}
The reasoning structure is segmented by special tokens such as So, Step, and Wait~\cite{bogdan2025thought}. 
We assume that activated neuron nodes in the middle layers trigger these special token expressions at the logit layers. At the same time, the neuron nodes themselves are activated by input tokens. To capture both effects, we treat logit-layer nodes as signals directed back to the neuron nodes by reversing their edge directions. We then apply the PageRank algorithm to identify the most influential nodes. We summarize the results in Figure~\ref{fig:token_data} and Figure~\ref{fig:token_model}, and the full details can be found in Appendix~\ref{app_sec:token_results}.

Figure~\ref{fig:token_data} presents the token distribution across datasets, highlighting the distinction between silent and activated tokens in LLMs when processing data from different domains. For example, numerical tokens such as 0 are more frequently activated in mathematical datasets, while general tokens like the period and ``the" appear across multiple domains. Furthermore, Figure~\ref{fig:token_model} illustrates that different models exhibit distinct activation preferences. After finetuning with distilled data, DS-Qwen-1.5B and DS-Llama-8B show increased activation of the token "**", suggesting that DeepSeek-distilled data encourages reliance on this token during generation. From a layer-level perspective, activated nodes consistently emerge in higher layers across both models and datasets, though not exclusively in the final layer; in the Qwen series, layers 22 to 27 frequently contain critical activated nodes.
\vspace{-5pt}
\paragraph{Case Study}

We further conduct case studies to illustrate how semantic merging and reasoning-trigger effects influence the reasoning process.
Based on the analyses in the previous sections, we observe that tokens, such as “the” and “.”, are identified as important by both analysis methods, although their functional roles vary across layers.
For instance, in Qwen3, these tokens tend to merge in middle layers (e.g., around layer 15), while in higher layers (e.g., beyond layer 25), they often act as reasoning triggers.
Accordingly, we design case studies to examine how the roles of such tokens evolve across layers during reasoning.


Specifically, we mute neurons in the LLM when they correspond to a given token at a specified layer. We show that muting neurons in middle layers leads to semantic changes, as illustrated in Figure~\ref{fig:case_study_semantic}. When the neuron node associated with “the” is deleted at layer 20, the English semantic pathway is disrupted, activating Chinese neurons that share similar semantic signals. As a result, the model produces output in Chinese, effectively translating from the original English text. 
In the Reasoning trigger case, deleting the neuron node “.” at layer 25 introduces a new reasoning token such as ``So", ultimately leading the model to produce the correct answer. The full results are shown in Appendix~\ref{app_sec:case_studies}.

\begin{figure*}
\vspace{-30pt}
    \centering
    \includegraphics[width=0.9\linewidth]{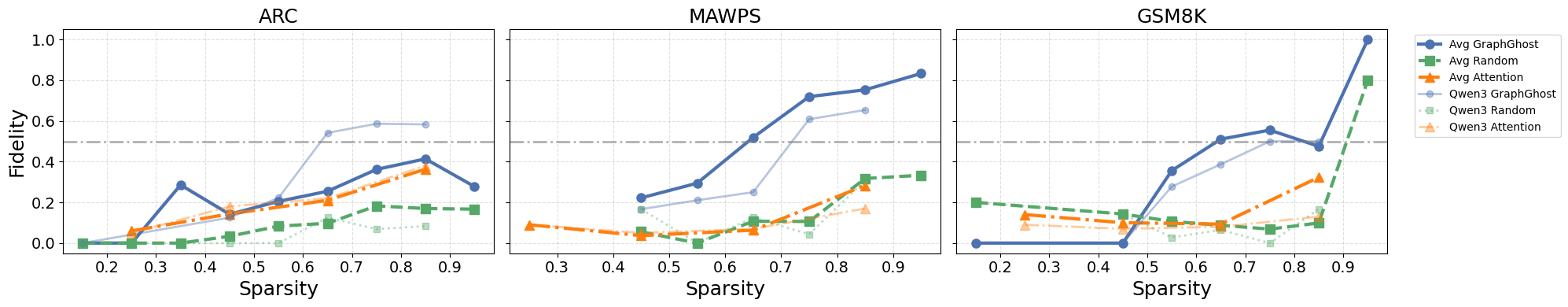}
     \vspace{-10pt}
    \caption{Sparsity Vs. Fidelity in various datasets and models.}
    \label{fig:s_f}
    \vspace{-14pt}
\end{figure*}
\vspace{-5pt}
\section{Quantitative Experiments}
\vspace{-5pt}
\label{sec:controlling}
To evaluate whether the structural patterns identified within LLMs are functionally relevant to reasoning behavior,
We perform quantitative experiments that examine whether GraphGhost-identified tokens exhibit consistent and measurable significance.
We first introduce the evaluation metrics in Section~\ref{sub_sec:metrics}.
The evaluation considers both the sample-level view (Section~\ref{sub_sec:pert_contribute})
and the dataset-level view (Section~\ref{sub_sec:pertubation}).

\vspace{-5pt}
\subsection{Metrics}
\label{sub_sec:metrics}
We conduct quantitative evaluations using Fidelity, InFidelity, and Sparsity metrics to assess whether model predictions remain consistent with the original, non-perturbed outputs~\cite{yeh2019fidelity}. In practice, we use gemini-2.5-flash-lite to determine whether two generated predictions are semantically identical. Sparsity measures the proportion of features identified as important~\cite{guidotti2018survey}. In the sample view, we evaluate contribution tokens using both Sparsity and Fidelity. Formally,
$
\mathrm{Sparsity} = \frac{|C|}{|T|}$, 
$
\mathrm{Fidelity}_T = \mathds{1}\!\left(LLM(C),\, LLM(T)\right),
$
where \(C\) denotes the set of contributing tokens, \(T\) denotes the full token set, and \(\mathds{1}(\cdot,\cdot)\) is an indicator function that returns 1 if the two outputs are identical and 0 otherwise. In the dataset view, we evaluate whether muting the identified nodes affects the model’s prediction and report InFidelity, defined as
$
\mathrm{InFidelity} = 1 - \mathrm{Fidelity}
= \mathds{1}\!\left(LLM(T),\, \hat{LLM}(T)\right),
$
which captures cases where muting the selected nodes in \(\hat{LLM}(\cdot)\) results in a change in the model’s output.



\vspace{-5pt}
\subsection{Contribution Token Selection in GraphGhost: Sample View}
\label{sub_sec:pert_contribute}

To evaluate whether GraphGhost can identify tokens that are functionally important for the reasoning process,
we compare it against attention-based explanation methods~\citep{clark2019doesbertlookat, liu2024understandingincontextlearningcontrastive} and random baselines.
Instead of analyzing individual tokens, we adapt the same pipeline as GraphGhost to evaluate its effectiveness. The detailed settings are provided in Appendix~\ref{app_sec:attention_details}. 

The comparative results are summarized in Figure~\ref{fig:s_f}. For a detailed breakdown of the performance for each individual model without averaging, we provide the complete results in Appendix~\ref{app_sec:full_results} (Figure~\ref{fig:app_line}). GraphGhost consistently outperforms both random and attention-based baselines. This advantage is most pronounced on mathematical reasoning tasks (MAWPS and GSM8K), where the attention baseline performs comparably to random noise, indicating its failure to capture essential reasoning structures. In contrast, GraphGhost maintains high fidelity even under low sparsity constraints.
Moreover, on the challenging ARC dataset, GraphGhost demonstrates superior robustness; notably, it is the only method that surpasses the 0.5 fidelity threshold on Qwen3. These results suggest that while attention primarily captures surface-level correlations, GraphGhost is able to trace deeper, multi-hop structural dependencies that are critical for complex reasoning.

\subsection{Behaviors Controlled by Critical Nodes in GraphGhost: Dataset View}
\label{sub_sec:pertubation}

To evaluate whether GraphGhost successfully identifies the most influential tokens in LLMs, we mute specific neuron nodes and observe the resulting changes in the outputs. Specifically, neuron nodes are muted by setting the corresponding positions of their layer–token pairs to zero during the reasoning process.
We evaluate the impact through both case studies and statistical analysis.

\begin{figure}[ht]
\vspace{-5pt}
    \centering
    \includegraphics[width=0.9\linewidth]{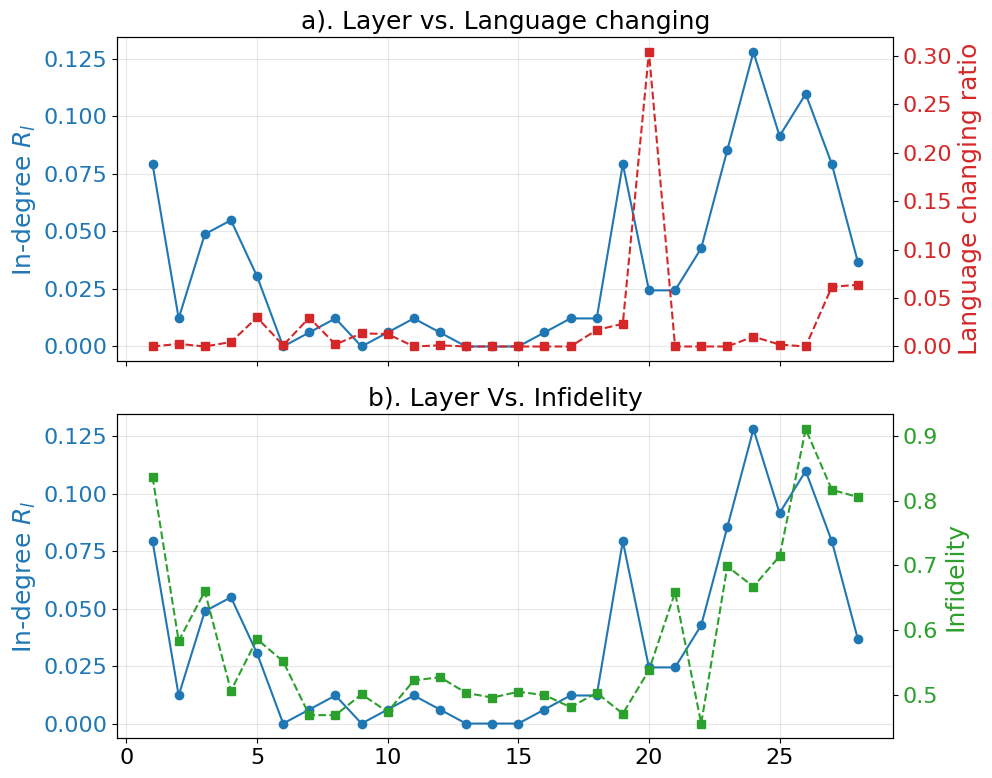}
     \vspace{-5pt}
    \caption{The impact of tokens on various layers. We mute the period token '.' for each layer.}
    \label{fig:layer_fig}
     \vspace{-12pt}
\end{figure}

\begin{table}[ht]

\resizebox{0.9\linewidth}{!}{
\begin{tabular}{lllll}
\toprule
\multirow{2}{*}{} & \multicolumn{2}{l}{Top1-10}                 & \multicolumn{2}{l}{Top11-20}                \\
                  & AVG    & MAX                                & AVG    & MAX                                \\
                  \midrule
Token             & -      & \textless{}space\textgreater{}\_27 & -      & 0\_23                              \\
Qwen3-0.6B        & 0.5845 & 0.9701                             & 0.3756 & 0.6958                             \\
Token             & -      & \textless{}space\textgreater{}\_27 & -      & \textless{}space\textgreater{}\_31 \\
Qwen2.5-1.5B      & 0.4586 & 0.9577                             & 0.3411 & 0.8956                             \\
Token             & -      & \textless{}space\textgreater{}\_31 & -      & \textless{}space\textgreater{}\_30 \\
Llama-8B          & 0.2808 & 0.8380                             & 0.2454 & 0.5028     \\
\bottomrule
\end{tabular}}
\caption{The impacts of top tokens. “MAX” denotes the maximum influential token among the top-K slots} 
\vspace{-10pt}
\label{tab:top_token}
\end{table}

We first analyze the relationship between the in-degree ratio across layers and the final reasoning predictions in the Qwen3-0.6B model, as illustrated in Figure~\ref{fig:layer_fig}. Neuron nodes with higher in-degree values are likely to merge semantic information from multiple sources. To verify this, we compute the language-changing ratio when muting individual neuron nodes across layers, defined as $R = \frac{\# \mathrm{non-English\ answer}}{\# \mathrm{All\ answer}}$, 
where a \textit{non-English answer} refers to any generated response that switches from English to another language after perturbation. As shown in Figure~\ref{fig:layer_fig}(a), the peaks of the language-changing ratio align closely with layers exhibiting high in-degree ratios in middle layers. Furthermore, as shown in Figure~\ref{fig:layer_fig}(b), Infidelity increases in the upper layers. This indicates that neuron nodes in higher layers are more likely to connect directly to logit neurons, thereby altering the underlying reasoning logic. This observation is consistent with the findings in Section~\ref{sec:graphghost_analysis}, where perturbations in middle layers primarily affect semantic expression, whereas perturbations in higher layers tend to modify the reasoning logic itself. 

We also evaluate the important tokens ranked by PageRank algorithm. We calculate the average InFidelity score when muting the corresponding neuron nodes and identify the maximum-impact neuron nodes within the top-$k$ positions. The results, summarized in Table~\ref{tab:top_token}, include the top-1–10 and top-11–20 nodes. 
Overall, the top-10 nodes show much higher influence, indicating that PageRank effectively identifies important nodes in GraphGhost. Larger models are more stable than smaller ones; Llama-8B achieves an InFidelity score of 0.28, whereas Qwen3 reaches 0.58. Notably, the <space> token is the most influential, especially in higher layers.





\vspace{-5pt}
\section{Conclusion}
\vspace{-5pt}
We present GraphGhost, a graph-based framework that reveals the internal structures underlying LLM reasoning. By linking token interactions and neuron activations to graph properties, GraphGhost shows how semantic merging and reasoning triggers shape model behavior. Our results demonstrate that LLM reasoning is governed by structured internal graphs, offering a principled path toward mechanistic interpretability.

\section*{Limitations}
The fidelity of GraphGhost relies heavily on the precision of the underlying attribution tracing method.
Because current attribution estimates can be noisy or incomplete, they may produce missing or spurious edges, yielding only partial views of the model’s reasoning process.
Advancing the design of circuit tracer would further enhance GraphGhost’s ability to identify key neuron nodes and capture more precise reasoning structures.

\section*{Potential Risk}
A potential risk of our method is that the extracted graph structures may be over-interpreted as definitive explanations of model behavior, while they only provide approximate views of internal information flow. We caution against such misuse and position our approach as a complementary analytical tool.
\bibliography{custom}
\appendix

\section{Appendix: Algorithm}
\label{app_sec:alg}
Here we present the algorithm of construct the sample and dataset view of GraphGhost in Algorithm~\ref{alg:graphghost_sample} and Algorithm~\ref{alg:graphghost_dataset}.
~

\begin{algorithm}[H]
\caption{GraphGhost Construction (Sample View)}
\label{alg:graphghost_sample}
\begin{algorithmic}[1]
    \State \textbf{Input:} A sample $s$ containing a final answer token $t_n$
    \State \textbf{Output:} 
        \begin{itemize}
            \item Sample-level GraphGhost $G^{\text{sample}} = (V, E)$ (unweighted)
            \item A local edge-weight map $W^{\text{local}}$ (for dataset aggregation only)
        \end{itemize}
    \State Initialize $V=\emptyset$, $E=\emptyset$, and $W^{\text{local}}(e)=0$ for all edges $e$
    \State Initialize frontier $T = \{t_n\}$
    \For{each token $t \in T$}
        \If{$t$ is not part of the question tokens}
            \State \textit{\# Step 1: Local logit-based attribution graph}
            \State $(\hat{T}, G_A') \gets \textsc{CircuitTracer}(t)$
            \Comment{$G_A' = (\hat{V}, \hat{E}, W^{\text{loc}})$}
            \State $T \gets T \cup \hat{T}$
            \State \textit{\# Step 2: Add edges to sample-level graph (structure only)}
            \State $V \gets V \cup \hat{V}$
            \State $E \gets E \cup \hat{E}$   \Comment{unweighted sample graph}
            \State \textit{\# Step 3: Accumulate local weights for dataset view}
            \For{each edge $e \in \hat{E}$}
                \State $W^{\text{local}}(e) \gets W^{\text{local}}(e) + W^{\text{loc}}(e)$
            \EndFor
        \EndIf
    \EndFor
    \State \Return $(G^{\text{sample}} = (V,E),\; W^{\text{local}})$
\end{algorithmic}
\end{algorithm}

\begin{algorithm}[H]
\caption{GraphGhost Construction (Dataset View)}
\label{alg:graphghost_dataset}
\begin{algorithmic}[1]
    \State \textbf{Input:} A set of samples $\mathcal{S} = \{s_1,\dots,s_m\}$
    \State \textbf{Output:} Dataset-level GraphGhost $G^{\text{data}} = (V^{\text{data}}, E^{\text{data}}, W^{\text{data}})$
    \State Initialize $V^{\text{data}}=\emptyset$, $E^{\text{data}}=\emptyset$, and $W^{\text{data}}(e)=0$
    \For{each sample $s_i \in \mathcal{S}$}
        \State $(G^{(i)}, W^{\text{local}}) \gets \textsc{GraphGhost-SampleView}(s_i)$
        \State $(V^{(i)}, E^{(i)}) \gets G^{(i)}$
        \State \textit{\# Merge sample graph structure}
        \State $V^{\text{data}} \gets V^{\text{data}} \cup V^{(i)}$
        \State $E^{\text{data}} \gets E^{\text{data}} \cup E^{(i)}$
        \State \textit{\# Accumulate edge weights across samples}
        \For{each edge $e \in E^{(i)}$}
            \State $W^{\text{data}}(e) \gets W^{\text{data}}(e) + W^{\text{local}}(e)$
        \EndFor
    \EndFor
    \State \Return $G^{\text{data}} = (V^{\text{data}}, E^{\text{data}}, W^{\text{data}})$
\end{algorithmic}
\end{algorithm}
\section{Appendix: Experiment setting}
\subsection{Synthetic Experiment}
\label{app_sec:sync_experiment}
We first generate directed graphs with 10 nodes and an edge density of 0.6, then randomly drop 60\% of the edges. From each resulting subgraph, we randomly select two nodes and retain the subgraph if the path length between them is 3. In total, we collect 100,000 such graph cases. For training, we use a 5-layer decoder-only transformer with a hidden size of 96 and 12 attention heads. The model is trained for 20,000 iterations to ensure stable convergence where the validation loss is the minimal.

For circuit tracing, we employ the ReLU version of the transcoder, with the dead feature window set to 50 and the L1 coefficient set to 0.0005. During visualization, we set the node ratio to 0.48 and the edge ratio to 0.86.

\subsection{Real-world Experiment}
\label{app_sec:real_world_exp}
The experimental settings for real-world data are summarized in Table~\ref{tab:real_world_setting}. We include both the thinking prompt “Let’s think step by step” and the ``\textless{}think\textgreater{}'' tag. For the transcoders, we select parameter configurations that converge to at least an MSE value of 0.001 at the first layer.
\begin{table}[ht]

\resizebox{\linewidth}{!}{\begin{tabular}{lllllll}
\toprule
\multicolumn{2}{l}{}                               & Qwen3-0.6B                                              & Qwen2.5-1.5B             & Llama-8B                                     & DS-Qwen                                            & DS-Llama                                             \\
\midrule
\multirow{6}{*}{Max Length}      & MAWPS (1772)    & 200                                                     & 200                      & 200                                                     & 400                                                     & 400                                                     \\
                                 & GSM8K (1319)    & 300                                                     & 300                      & 300                                                     & 500                                                     & 500                                                     \\
                                 & ProntoQA (1500) & 900                                                     & 900                      & 900                                                     & 900                                                     & 900                                                     \\
                                 & BoolQA (3270)   & 500                                                     & 500                      & 500                                                     & 500                                                     & 500                                                     \\
                                 & ARC (2376)      & 300                                                     & 300                      & 300                                                     & 500                                                     & 500                                                     \\
                                 & QASC (926)      & 300                                                     & 300                      & 300                                                     & 500                                                     & 500                                                     \\
                                 \midrule
\multirow{3}{*}{Thinking prompt} & Math            & Let's...                               & Let's... &Let's...                                & Let's... \textless{}\textgreater{} & Let's...\textless{}\textgreater{} \\
                                 & Logic           & Let's... \textless{}\textgreater{} & Let's... & Let's... \textless{}\textgreater{} & Let's... \textless{}\textgreater{} & Let's... \textless{}\textgreater{} \\
                                 & Science         & Let's... \textless{}\textgreater{} & Let's... & Let's... \textless{}\textgreater{} & Let's...\textless{}\textgreater{} & Let's... \textless{}\textgreater{} \\
                                 \midrule
\multirow{3}{*}{Transcoder}      & Edge Ratio      & 0.4                                                     & 0.4                      & 0.8                                                     & 0.4                                                     & 0.9                                                     \\
                                 & Node Ratio      & 0.5                                                     & 0.5                      & 0.9                                                     & 0.5                                                     & 0.9                                                     \\
                                 & L1\_coefficient & 0.0005                                                  & 0.0005                   & 0.00005                                                 & 0.0005                                                  & 0.00005                                                 \\
                                 \midrule
\multirow{6}{*}{Training Tokens} & MAWPS           & 5*10\textasciicircum{}6                                 & 5*10\textasciicircum{}6  & 5*10\textasciicircum{}6                                 & 5*10\textasciicircum{}6                                 & 5*10\textasciicircum{}6                                 \\
                                 & GSM8K           & 2*10\textasciicircum{}7                                 & 2*10\textasciicircum{}7  & 2*10\textasciicircum{}7                                 & 2*10\textasciicircum{}7                                 & 2*10\textasciicircum{}7                                 \\
                                 & ProntoQA        & 1*10\textasciicircum{}7                                 & 1*10\textasciicircum{}7  & 1*10\textasciicircum{}7                                 & 1*10\textasciicircum{}7                                 & 1*10\textasciicircum{}7                                 \\
                                 & BoolQA          & 5*10\textasciicircum{}6                                 & 5*10\textasciicircum{}6  & 5*10\textasciicircum{}6                                 & 2*10\textasciicircum{}7                                 & 2*10\textasciicircum{}7                                 \\
                                 & ARC             & 2*10\textasciicircum{}7                                 & 2*10\textasciicircum{}7  & 2*10\textasciicircum{}7                                 & 2*10\textasciicircum{}7                                 & 2*10\textasciicircum{}7                                 \\
                                 & QASC            & 2*10\textasciicircum{}7                                 & 2*10\textasciicircum{}7  & 2*10\textasciicircum{}7                                 & 2*10\textasciicircum{}7                                 & 2*10\textasciicircum{}7         \\
                                 \bottomrule
\end{tabular}}

\caption{The experiment setting for real world cases. ``Let's...'' denotes the prompt ``Let's think step-by-step'', and ``\textless{}\textgreater{}'' denotes ``\textless{}think\textgreater{}''}
\label{tab:real_world_setting}

\end{table}
\subsection{Case studies}
\label{app_sec:case_studies}
We show the full case studies of how neuron nodes change the LLMs behaviors in Figure~\ref{fig:case_study_reasoning_structure_full} and Figure~\ref{fig:case_study_semantic_full}.
 We show that muting neurons in middle layers leads to semantic changes, as illustrated in Figure~\ref{fig:case_study_semantic_full}. When the neuron node associated with “the” is deleted at layer 20, the English semantic pathway is disrupted, activating Chinese neurons that share similar semantic signals. As a result, the model produces output in Chinese, effectively translating from the original English text. Furthermore, in the word-expression change case, deleting the neuron associated with “.” at layer 17 causes the token “/” to replace “÷”.

In Figure~\ref{fig:case_study_reasoning_structure_full}, we demonstrate how GraphGhost influences reasoning by altering key steps or driving the generation into loops. In the loop case, deleting the neuron node “the” at layer 27 causes the reasoning to degenerate into a repetition, producing “theShaper Information.” These cases suggest that even a single neuron in the middle layer can fundamentally change the outcome of the reasoning process. In the key-step change case, deleting the neuron node “.” at layer 25 introduces a new reasoning token such as ``So", ultimately leading the model to produce the correct answer. 
\begin{figure*}
\centering
\resizebox{0.9\textwidth}{!}{%
  \begin{minipage}{1.4\textwidth}
    \begin{tcolorbox}[title=Change the semantic, colback=white, colframe=black!50]
      \textbf{Change the language}\par

      \textit{Question:} {A ) nervous and digestive B ) nervous and circulatory C ) respiratory and digestive D ) respiratory and circulatory  Question: Which two body systems would most directly remove extra fluid from a person's lungs?}\par
        
      \textit{Original answer:} {... B) Nervous and circulatory systems: The nervous system is responsible for regulating the body's response to stress, while the circulatory system is responsible for transporting blood throughout the body. Neither of these systems directly remove extra fluid from the lungs...}\par
      \textbf{\textit{Mute the neuron node 'the' at layer 20}}\par
      \textit{Perturbed answer:} {... B) Nervous and circulatory systems: The nervous system is responsible for controlling the - circulatory system is responsible for transporting blood throughout the\begin{CJK}{UTF8}{gbsn}分区肺\end{CJK}. While the\begin{CJK}{UTF8}{gbsn}分区肺需要血液来清除多余的水分，这些系统本身并不直接参与这个过程。\end{CJK}...}\par
    \noindent\rule{\linewidth}{0.2pt}
      \textbf{Change the word expression}\par
      %

      \textit{Question:} {To determine how many trips Jerry will make, we need to calculate the total number of trays he needs to pick up and then divide by the number of trays he can carry at a time.}\par
      \textit{Original answer:} {... we divide the total number of trays, 16.0 trays ÷ 8.0 trays per trip = 2 trips...}\par
\textbf{\textit{Mute the neuron node '.' at layer 17}}\par
      \textit{Perturbed answer:} {... Number of trips = Total trays / 8 = 16 / 8 = 2 trips ...}\par


    \end{tcolorbox}
  \end{minipage}
}
\caption{Semantic control in middle layers modulates changes in semantic representations. In the language-switching case, the generated characters are direct translations of the original answer.} 
\label{fig:case_study_semantic_full}
\end{figure*}
\begin{figure*}[!t]
\centering
\resizebox{0.9\textwidth}{!}{%
  \begin{minipage}{1.4\textwidth}
    \begin{tcolorbox}[title=Change the reasoning structure, colback=white, colframe=black!50]

    \textbf{Failed into loop}\par
      %

      \textit{Question:} {Alyssa picked 25.0 limes and Mike picked 32.0 limes. Tom picked 12.0 plums. How many limes were picked in all?
Let's think step by step}\par
      \textit{Original answer:} {Step 1: Identify the number of limes picked by Alyssa and Mike.
Alyssa picked 25.0 limes.
Mike picked 32.0 limes. ...}\par
\textbf{\textit{Mute the neuron node 'the' at layer 27}}\par
      \textit{Perturbed answer:} {Step 1: Identify theSharper Information
In this problem, we are given theSharper Information about theSharper Information about Alyssa, Mike, and Tom's limes and plums. We need to find theSharper Information about theSharper Information about theSharper ...}\par

 \noindent\rule{\linewidth}{0.2pt}

      \textbf{Change the key steps}\par

      \textit{Question:} {There are 7.0 crayons in the drawer and 6.0 crayons on the desk. Sam placed 4.0 crayons and 8.0 scissors on the desk. How many crayons are now there in total?
Let's think step by step (ground truth: 17)}\par
        
      \textit{Original answer:} {... - Sam placed 4.0 crayons on the desk.
- Sam also placed 8.0 scissors on the desk. .... - Total crayons on the desk = 7.0 + 4.0 + 8.0
- Total crayons on the desk = 19.0}\par
      \textbf{\textit{Mute the neuron node '.' at layer 25}}\par
      \textit{Perturbed answer:} {... - Sam placed 4.0 crayons on the desk.
- So, the total number of crayons on the desk is now 6.0 (initial crayons) + 4.0 (Sam's crayons) = 10.0 crayons. ... - So, the total number of crayons = 7.0 (drawer) + 10.0 (desk) = 17.0 crayons.}\par
       
      


    \end{tcolorbox}
  \end{minipage}
}
\caption{Reasoning triggers changes on the reasoning structures}
\label{fig:case_study_reasoning_structure_full}
\end{figure*}
\subsection{Attention-based Baseline Implementation}
\label{app_sec:attention_details}

To adapt the standard attention analysis~\citep{clark2019doesbertlookat} for reasoning tasks, we implement a specific Answer-to-Input saliency measure. We first calculate the mean attention matrix $\bar{\mathbf{A}}$ by averaging weights across all heads in the final layer. The importance score $S_i$ for each input token $i$ is then derived by accumulating the attention it receives from the entire generated answer sequence: 
\[
S_i = \sum_{j \in \text{Answer}} \bar{\mathbf{A}}_{j, i}
\]

This score represents the total attention mass focused on input $i$ during the reasoning generation. Input tokens are ranked by $S_i$, and the top-$k\%$ are retained to construct the perturbed prompt for fidelity evaluation, following the same protocol as GraphGhost.

\section{GraphGhost Results}
\label{app_sec:token_results}
In this section, we present the specific tokens ranked by graph algorithms. For the semantic merging case, where tokens are sorted by in-degree, Table~\ref{tab:sem_datasets} and Table~\ref{tab:sem_model} list the top tokens across different datasets and models. Table~\ref{tab:merged_semantic} highlights significant tokens that are merged across multiple layers of LLMs. For logic shifting, we report the top tokens for models and datasets in Table~\ref{tab:pagerank_models} and Table~\ref{tab:pagerank_datasets}, respectively.

\begin{table}[ht]

\resizebox{\linewidth}{!}{\begin{tabular}{rllllll}
\toprule
\multicolumn{1}{l}{\multirow{2}{*}{Top tokens}} & \multicolumn{6}{l}{Dataset}                                                                                                                                                                            \\
\multicolumn{1}{l}{}                           & MAWPS                               & GSM8K                               & ARC                                 & QASC     & ProntoQA                            & BoolQA                              \\
\midrule
1                                              & .                                   & the                                 & the                                 & ,        & .                                   & the                                 \\
2                                              & the                                 & .                                   & ,                                   & the      & uses                                & ,                                   \\
3                                              & \textless{}space\textgreater{}      & \textless{}space\textgreater{}      & is                                  & is       & .\textbackslash{}n                  & .                                   \\
4                                              & 0                                   & ,                                   & .                                   & .        & ump                                 & is                                  \\
5                                              & :                                   & 2                                   & are                                 & are      & is                                  & a                                   \\
6                                              & 1                                   & .\textbackslash{}n\textbackslash{}n & that                                & that     & the                                 & that                                \\
7                                              & .\textbackslash{}n                  & 1                                   & us                                  & are      & uses                                & are                                 \\
8                                              & is                                  & \$                                  & .\textbackslash{}n\textbackslash{}n & of       & .\textbackslash{}n\textbackslash{}n & in                                  \\
9                                              & ,                                   & 0                                   & 1                                   & to       & 1                                   & .\textbackslash{}n\textbackslash{}n \\
10                                             & .\textbackslash{}n\textbackslash{}n & is                                  & given                               & directly & given                               & was     \\                           
\bottomrule
\end{tabular}}
\caption{The top tokens sort by the in-degree values across the datasets}
\label{tab:sem_datasets}

\end{table}
\begin{table}[ht]

\resizebox{\linewidth}{!}{\begin{tabular}{llllll}
\toprule
\multirow{2}{*}{Top token} & \multicolumn{5}{l}{Model}                                                                                                                                                         \\
                           & Qwen3-0.6B                          & Qwen2.5-1.5B                   & DS-Qwen-1.5B                        & Llama-8B-Instruct              & DS-Llama-8B                         \\
                           \midrule
1                          & .                                   & \textless{}space\textgreater{} & \textless{}space\textgreater{}      & \textless{}space\textgreater{} & contin                              \\
2                          & the                                 & .                              & ,                                   & .                              & \textless{}space\textgreater{}      \\
3                          & \textless{}space\textgreater{}      & the                            & .                                   & the                            & **                                  \\
4                          & 0                                   & =                              & the                                 & ,                              & .                                   \\
5                          & :                                   & .\textbackslash{}n             & :                                   & :                              & ,                                   \\
6                          & 1                                   & ,                              & .\textbackslash{}n\textbackslash{}n & =                              & .\textbackslash{}n\textbackslash{}n \\
7                          & .\textbackslash{}n                  & 1                              & to                                  & to                             & i                                   \\
8                          & is                                  & has                            & is                                  & 0                              & **                                  \\
9                          & ,                                   & had                            & =                                   & of                             & the                                 \\
10                         & .\textbackslash{}n\textbackslash{}n & \textbackslash{}n              & 0                                   & and                            & \textbackslash{}\textbackslash{}   \\
\bottomrule
\end{tabular}}
\caption{The top tokens sort by the in-degree values across the models}
\label{tab:sem_model}

\end{table}

\begin{table}[ht]

\resizebox{\linewidth}{!}{\begin{tabular}{llllll}
\toprule
\multicolumn{1}{l}{Top nodes} & Qwen3-0.6B                             & Qwen2.5-1.5B           & DS-Qwen-1.5B                            & Llama-8B-Instruct                  & DS-Llama-8B                              \\
\midrule
1                             & 0\_27                                  & :\textbackslash{}n\_27 & .\_27                                   & \textbackslash \_31                & \{\_31                                   \\
2                             & \textbackslash \_27                    & :\_27                  & the\_26                                 & .\_31                              & .\textbackslash{}n\textbackslash{}n\_31  \\
3                             & .\_27                                  & :\textbackslash{}n\_26 & the\_27                                 & :\_31                              & \textbackslash \_31                      \\
4                             & the\_26                                & :\_26                  & ,\_26                                   & 0\_31                              & \textbackslash **\_31                    \\
5                             & :\_27                                  & \textbackslash \_27    & ,\_27                                   & the\_31                            & .\_31                                    \\
6                             & :\_26                                  & .\_27                  & .\_26                                   & more\_31                           & .\textbackslash{}n\_31                   \\
7                             & the\_22                                & 0\_27                  & **\_27                                  & ,\_31                              & **\_31                                   \\
8                             & .\_24                                  & .\textbackslash{}n\_27 & the\_25                                 & =\_31                              & \textbackslash{}n\textbackslash{}n\_31   \\
9                             & 0\_26                                  & \textbackslash .\_27   & ,\_25                                   & \textbackslash \textbackslash \_31 & **\textbackslash{}n\textbackslash{}n\_31 \\
10                            & .\_26                                  & \textbackslash \_25    & 0\_27                                   & the\_30                            & :\textbackslash{}n\_31                   \\
11                            & of\_26                                 & :\_25                  & to\_27                                  & ,\_30                              & \textbackslash \textbackslash{}n\_30     \\
12                            & 0\_23                                  & the\_26                & a\_27                                   & of\_31                             & \textbackslash{}\_31                     \\
13                            & .\_22                                  & .\_26                  & .\_25                                   & :\_30                              & vase\_0                                  \\
14                            & :\_24                                  & \textbackslash \_26    & the\_22                                 & is\_31                             & ,\_31                                    \\
15                            & :\_25                                  & the\_25                & .\textbackslash{}n\textbackslash{}n\_27 & 0\_30                              & \textbackslash{}n\_31                    \\
16                            & \textbackslash{}n\textbackslash{}n\_27 & .\_25                  & **\_26                                  & of\_30                             & i\_29                                    \\
17                            & \textbackslash{}n\textbackslash{}n\_26 & the\_23                & \textbackslash \_27                     & \textbackslash \_30                & contin\_31                               \\
18                            & :\_22                                  & \textbackslash .\_26   & the\_21                                 & to\_31                             & contin\_29                               \\
19                            & .\_23                                  & :\textbackslash{}n\_25 & I\_27                                   & pages\_31                          & contin\_30                               \\
20                            & the\_23                                & the\_22                & to\_26                                  & up\_31                             & **\textbackslash{}n\_31  \\
\bottomrule
\end{tabular}}
\caption{Top tokens ranked by PageRank across LLMs}
\label{tab:pagerank_models}

\end{table}
\begin{table}[ht]

\resizebox{\linewidth}{!}{\begin{tabular}{lllllll}
\toprule
\multicolumn{1}{l}{Top tokens} & MAWPS                                  & GSM8K               & BoolQA  & ProntoQA               & SCQA    & ARC     \\
\midrule
1                              & 0\_27                                  & the\_27             & the\_27 & .\_26                  & the\_27 & the\_27 \\
2                              & \textbackslash \_27                    & 0\_27               & the\_26 & is\_27                 & .\_27   & a\_27   \\
3                              & .\_27                                  & .\_27               & ,\_27   & .\textbackslash{}n\_27 & ,\_27   & .\_27   \\
4                              & the\_26                                & \textbackslash \_27 & a\_26   & .\_25                  & ,\_26   & the\_26 \\
5                              & :\_27                                  & the\_26             & a\_27   & ,\_25                  & are\_27 & a\_26   \\
6                              & :\_26                                  & :\_27               & .\_27   & ,\_26                  & is\_27  & the\_22 \\
7                              & the\_22                                & **\_27              & the\_22 & 1\_27                  & are\_26 & the\_25 \\
8                              & .\_24                                  & of\_26              & the\_25 & .\textbackslash{}n\_22 & a\_27   & a\_23   \\
9                              & 0\_26                                  & ,\_27               & the\_23 & ump\_27                & a\_26   & ,\_27   \\
10                             & .\_26                                  & .\_24               & ,\_26   & 1\_26                  & the\_25 & the\_21 \\
11                             & of\_26                                 & to\_26              & the\_21 & .\_22                  & )\_27   & is\_27  \\
12                             & 0\_23                                  & the\_25             & is\_27  & to\_25                 & are\_22 & the\_23 \\
13                             & .\_22                                  & ,\_26               & ,\_22   & -\_26                  & .\_25   & a\_25   \\
14                             & :\_24                                  & .\_26               & ,\_21   & given\_23              & the\_22 & a\_22   \\
15                             & :\_25                                  & (\_27               & .\_24   & the\_26                & )\_26   & a\_24   \\
16                             & \textbackslash{}n\textbackslash{}n\_27 & the\_23             & ,\_20   & the\_25                & "\_27   & a\_21   \\
17                             & \textbackslash{}n\textbackslash{}n\_26 & 0\_26               & a\_25   & is\_26                 & ,\_25   & to\_27  \\
18                             & :\_22                                  & the\_22             & a\_23   & \textbackslash \_25    & to\_26  & and\_27 \\
19                             & .\_23                                  & to\_25              & ,\_25   & ster\_27               & are\_24 & .\_24   \\
20                             & the\_23                                & ,\_25               & a\_22   & the\_22                & of\_27  & the\_24\\
\bottomrule
\end{tabular}}
\caption{Top tokens ranked by PageRank across datasets}
\label{tab:pagerank_datasets}
\end{table}
\begin{table*}[ht]

\resizebox{\textwidth}{!}{
\begin{tabular}{l|ll|rr|ll}
\toprule
Model                         & Dataset                 & Select Token         & \multicolumn{1}{l}{Layers} & \multicolumn{1}{l}{\# In-degree} & New tokens                             & Description                                        \\
\midrule
\multirow{16}{*}{Qwen3-0.6B}  & \multirow{4}{*}{ARC}    & \multirow{4}{*}{the} & 1                          & 26                               & is, not, of, in, and                   & Basic function words                               \\
                              &                         &                      & 15                         & 141                              & Because, But, could, bird              & Logical connectors/conditionals + noun             \\
                              &                         &                      & 20                         & 289                              & , Wait, DNA, .\textbackslash{}n        & Reasoning/pausing words + proper nouns             \\
                              &                         &                      & 25                         & 481                              & -check, A, 2), Carboon                 & answer-related symbols and entities                \\
                              
                              & \multirow{12}{*}{MAWPS} & \multirow{4}{*}{the} & 1                          & \multicolumn{3}{l}{None neurons}                                                                                               \\
                              &                         &                      & 15                         & 40                               & Calculate, add, in                     & Math calculation actions                           \\
                              &                         &                      & 20                         & 100                              & 1, \$, =, Answer, Step                 & Answer template with math symbols                  \\
                              &                         &                      & 25                         & 143                              & .\textbackslash{}n, how, think         & Reasoning/reflective tokens                        \\
                              &                         & \multirow{4}{*}{.}   & 1                          & 2                                & ., :\textbackslash{}n\textbackslash{}n & Sentence boundary markers                          \\
                              &                         &                      & 15                         & 14                               & Question, Answer, Step                 & Question–answer framing tokens                     \\
                              &                         &                      & 20                         & 134                              & \$, 2, + divide                        & Mathematical symbols and operations                \\
                              &                         &                      & 25                         & 147                              & How many, spent, together              & Math word-problem phrasing tokens                  \\
                              &                         & \multirow{4}{*}{,}   & 1                          & 3                                & First, Next                            & Sequencing/transition words                        \\
                              &                         &                      & 15                         & 1                                & ,                                      & Basic punctuation                                  \\
                              &                         &                      & 20                         & 11                               & Since, To                              & Ttransition words                                  \\
                              &                         &                      & 25                         & 30                               & How, total                             & Question words focusing on totals or amounts       \\
\multirow{8}{*}{Qwen2.5-1.5B} & \multirow{8}{*}{MAWPS}  & \multirow{4}{*}{the} & 1                          & 32                               & Calculate, add, of, in                 & Function words + common math verbs                 \\

\midrule
                              &                         &                      & 15                         & \multicolumn{3}{l}{None neurons}                                                                                               \\
                              &                         &                      & 20                         & 38                               & Multiply, Question                     & Math action words                                  \\
                              &                         &                      & 25                         & 54                               & 1, Answer, step                        & Answer-step related tokens                         \\
                              &                         & \multirow{4}{*}{.}   & 1                          & 16                               & \textbackslash{}n, 1                   & Formatting symbols and numeric markers             \\
                              &                         &                      & 15                         & 1                                & .                                      & Basic punctuation                                  \\
                              &                         &                      & 20                         & 49                               & =,+, got, Answer, Question             & Math operators mixed with answer framing tokens    \\
                              &                         &                      & 25                         & 63                               & \$, *                                  & Mathematical symbols                               \\
\multirow{8}{*}{Llama-8BI}    & \multirow{8}{*}{MAWPS}  & \multirow{4}{*}{the} & 1                          & 7                                & Question, For                          & Question framing / introductory tokens             \\
\midrule
                              &                         &                      & 2                          & 18                               & of, to, ing                            & Basic function words                               \\
                              &                         &                      & 30                         & 310                              & 2, \textbackslash{}n, If, eggs         & Numbers, line breaks, conditional phrasing         \\
                              &                         &                      & 31                         & 313                              & =, add, or                             & Mathematical operators and logical connectors      \\
                              &                         & \multirow{4}{*}{.}   & 1                          & 56                               & 1, yen, Question, to, ing              & Numeric units and question words                   \\
                              &                         &                      & 2                          & 72                               & all, and, company                      & Conjunctions and basic nouns                       \\
                              &                         &                      & 30                         & 210                              & 2, \textbackslash{}n, If, eggs         & Numbers with conditionals and real-world nouns     \\
                              &                         &                      & 31                         & 574                              & =+, 111, dogs                          & Numbers combined with operators and concrete nouns \\
                              \bottomrule
\end{tabular}
}
\caption{The examples of how LLMs merge the semantics.}
\label{tab:merged_semantic}

\end{table*}

\section{Additional Quantitative Results}
\label{app_sec:full_results}

\begin{figure*}[ht]
    \centering
    \includegraphics[width=\textwidth]{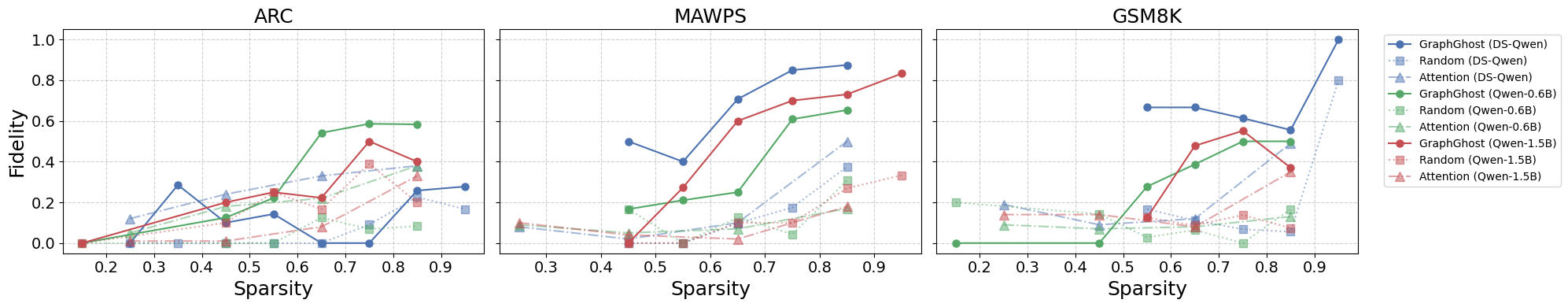} 
    \caption{Detailed fidelity-sparsity curves for individual models across ARC, MAWPS, and GSM8K datasets. This figure shows the non-averaged performance for GraphGhost compared to Random and Attention baselines.}
    \label{fig:app_line}
\end{figure*}

Figure~\ref{fig:app_line} presents the detailed fidelity-sparsity curves for all evaluated models (DS-Qwen-1.5B, Qwen3-0.6B, and Qwen2.5-1.5B) across the three datasets. Unlike the aggregated results in the main text (Figure~\ref{fig:s_f}), this figure illustrates the individual performance trajectories for each model without averaging.

\end{document}